\documentclass[letterpaper, 10 pt, conference]{ieeeconf}  
\usepackage{subcaption}
\usepackage{graphicx}  
\usepackage{float}
\usepackage{amsmath} 
\usepackage{amssymb} 
\usepackage{stfloats} 
\usepackage{textcomp}
\usepackage{booktabs} 
\usepackage{marvosym}
\usepackage{multirow}
\usepackage{caption}
\usepackage{url}
\usepackage{placeins}
\usepackage{afterpage}
\IEEEoverridecommandlockouts                              

\usepackage[dvipsnames]{xcolor}

\title{\LARGE \bf
Hysteresis-Aware Neural Network Modeling and Whole-Body Reinforcement Learning Control of Soft Robots
}

\author{Zongyuan Chen$^{1*}$, Yan Xia$^{1*}$, Jiayuan Liu$^{1}$, Jijia Liu$^{1}$, Wenhao Tang$^{14}$, Jiayu Chen$^{1}$, Feng Gao$^{1}$, \\Longfei Ma$^{1}$, Hongen Liao$^{1,2}$, Yu Wang$^{1}$, Chao Yu$^{1}\textsuperscript{\Letter}$, Boyu Zhang$^{2}\textsuperscript{\Letter}$, Fei Xing$^{1}\textsuperscript{\Letter}$ 
\thanks{$^{*}$Equal Contributions.}
\thanks{{\Letter} Corresponding Authors.}
\thanks{$^{1}$Tsinghua University, Beijing, 100084, China.
        {\tt\small chenzong23@mails.tsinghua.edu.cn}}   
\thanks{$^{2}$Shanghai Jiao Tong University, Shanghai, 200030, China.}%
\thanks{$^{3}$Tsinghua Shenzhen International Graduate School,
Shenzhen, 518055, China.}
}

\begin{document}
\maketitle
\thispagestyle{empty}
\pagestyle{empty}

\begin{abstract}
Soft robots exhibit inherent compliance and safety, which makes them particularly suitable for applications requiring physical interaction with humans, such as surgery procedures. However,  their nonlinear and hysteretic behavior, resulting from the properties of soft materials, presents substantial challenges for accurate modeling and control.
In this study, we present a soft robotic system designed for surgical applications and propose a hysteresis-aware whole-body neural network model that accurately captures and predicts the soft robot’s whole-body motion, including its hysteretic behavior. Building upon the high-precision dynamic model, we construct a highly parallel simulation environment for soft robot control and apply an on-policy reinforcement learning algorithm to efficiently train whole-body motion control strategies.
Based on the trained control policy, we developed a soft robotic system for surgical applications and validated it through phantom-based laser ablation experiments in a physical environment. The results demonstrate that the hysteresis-aware modeling reduces the Mean Squared Error (MSE) by 84.95\% compared to traditional modeling methods. The deployed control algorithm achieved a trajectory tracking error ranging from 0.126 to 0.250 mm on the real soft robot, highlighting its precision in real-world conditions. The proposed method showed strong performance in phantom-based surgical experiments, demonstrates its potential for complex scenarios, including future real-world clinical applications.

\end{abstract}


\section{INTRODUCTION}
Soft robots are typically constructed from soft material that extends, bends, and twists according to the actuation provided by air pressure or cables \cite{piqueControllingSoftRobotic2022}. They have a continuously deformable structure offer enhanced flexibility, compliance, and a high degree of freedom \cite{cakurdaDeepLearningMethods2024,terryn2017self,jiao2019vacuum}. In recent years, soft pneumatic robots have been widely introduced into tasks involving interaction with the human body, especially in medical and surgical applications \cite{wuDeepLearningBasedCompliantMotion2022}. Compared with conventional rigid surgical instruments, soft robots can provide more degrees of freedom to navigate easily through the complex and tortuous structures in the abdominal cavity. They also have good biocompatibility with patient tissues, reducing the risk of tissue damage and complications \cite{zhangLaserEndoscopicManipulator2021}.

However, soft robots are morphologically diverse, and their dynamics is highly influenced by material and structural design. 
Modeling soft robots presents significant challenges due to the intrinsic material nonlinearity~\cite{webster2010design} and hysteresis properties. The difficulty in obtaining accurate models for precise control increases the risk of failure or damage in applications. Therefore, achieving accurate modeling and control of soft robots remains a critical and challenging research problem.
\begin{figure}[h]
    \centering
    \includegraphics[width=1.0\linewidth]{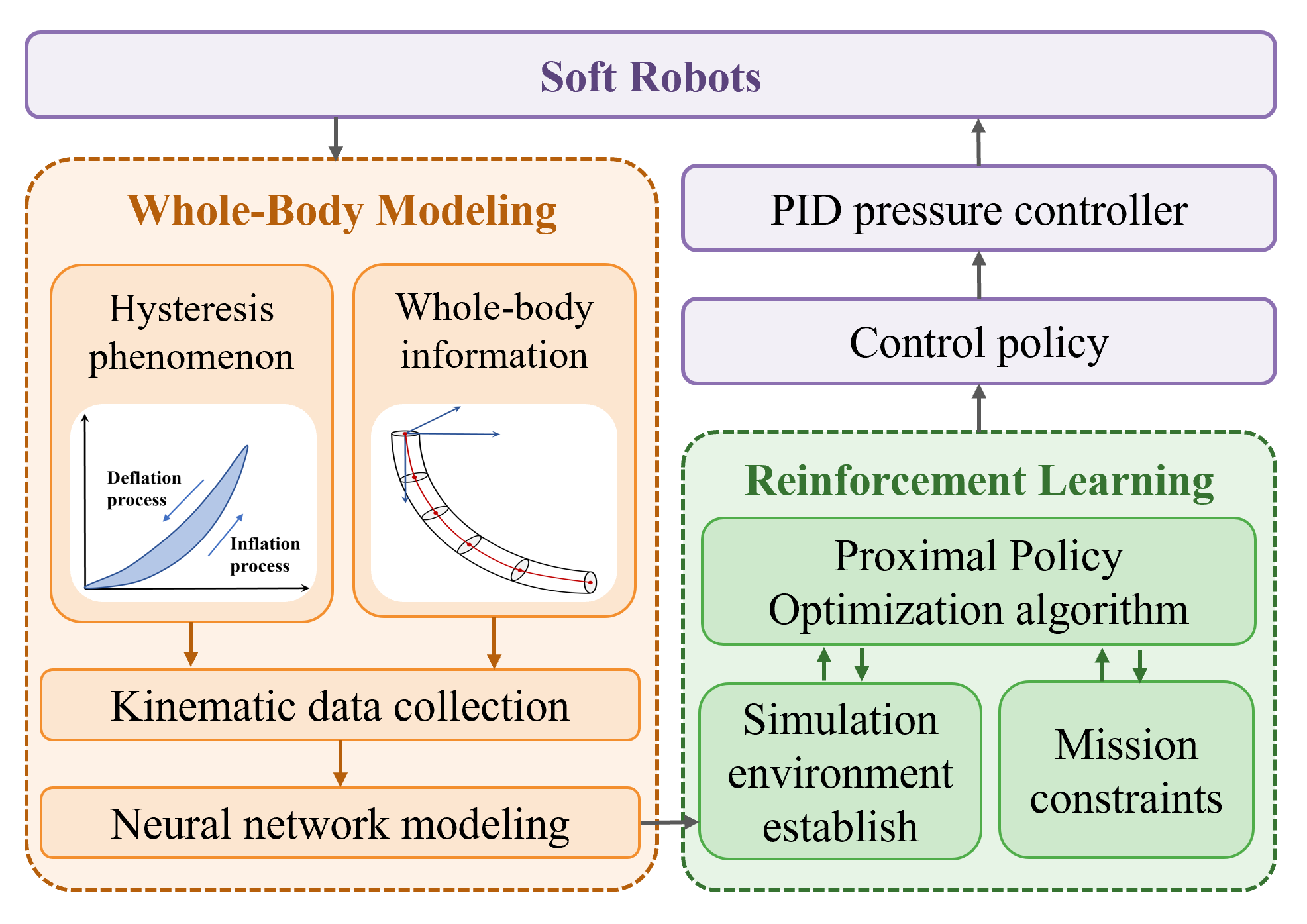}
    \caption{The proposed framework combining hysteresis-aware whole-body modeling and reinforcement learning for soft robot control.}
    \label{fig:fig1 general framework}
\end{figure}

\begin{figure*}[!b]
    \centering
    \includegraphics[width=0.9\linewidth]{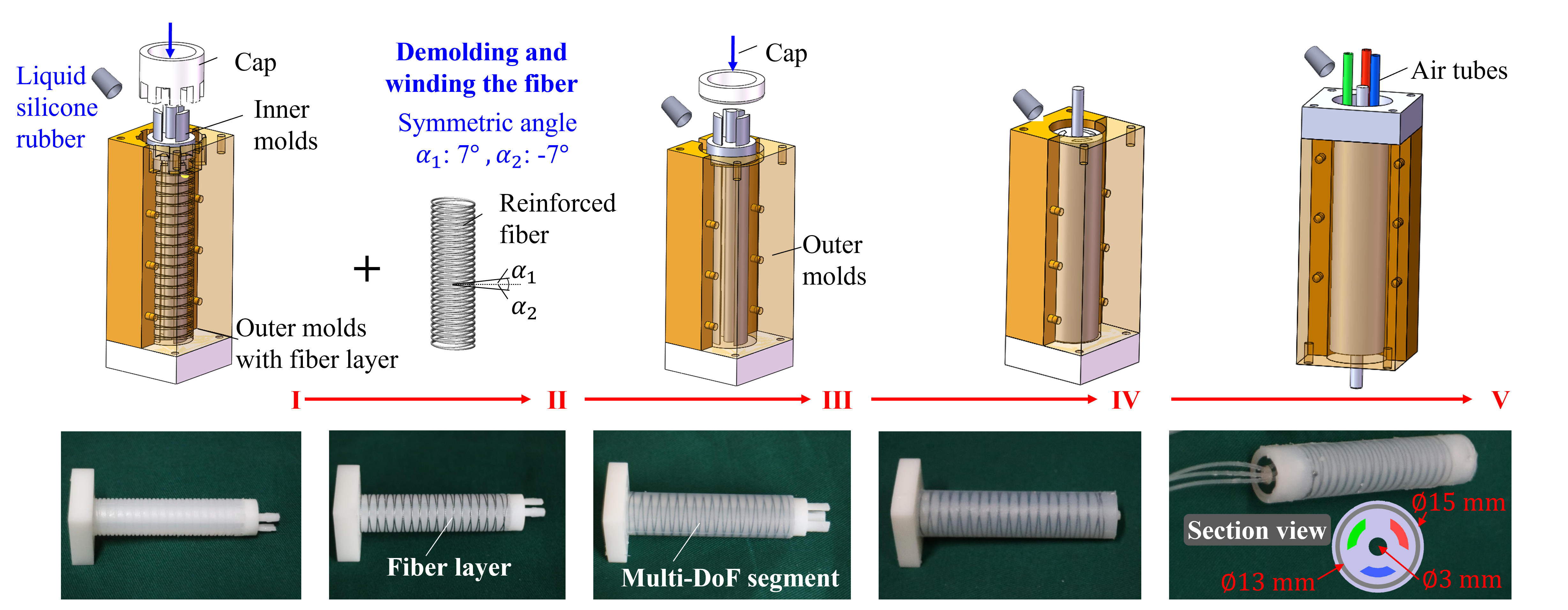}
    \caption{Manufacturing process of a fiber-constrained three-cavity pneumatic soft robot.}
    \label{fig:fabrication}
\end{figure*}

To model soft robots, conventional methods are mainly based on Piecewise Constant-Curvature (PCC) kinematic models \cite{zhou2022modeling} and Cosserat rod theory \cite{jones2006practical}. These methods are widely used for modeling soft robots with simple materials and structures. 
However, these methods rely on several simplifying assumptions, such as material homogeneity, pressure chambers maintaining constant circumferential geometry during actuation, and the independence of multiple chambers. These assumptions are rarely met in practical soft robotic systems. As application scenarios broaden, soft-bodied robots are increasingly integrated with functional attachments, further limiting the applicability of traditional analytical modeling methods. In recent years, data-driven methodologies have been applied for modeling flexible robots \cite{nazeerRLBasedAdaptiveController2024}, \cite{zhengVisionBasedOnlineKey2024a}. Specifically, multi-layer perceptrons (MLPs) have demonstrated superior performance, attributable to their capacity for non-linear fitting and feature characterization \cite{louControllingSoftRobotic2024}. 

The motion of soft robots exhibits hysteresis, indicating that their shape variables are influenced not only by the current inputs but also by the history of the input trajectories. As a result, identical inputs can produce different outputs during forward and backward motions, constituting one of the primary sources of modeling error \cite{wu2020model}. Previous studies have used recurrent neural networks (RNNs) to process sequential data and convolutional neural networks (CNNs) to model spatially rearranged pressure inputs. The study in \cite{zhangNovelHysteresisModelling2019} modeled hysteresis by inferring the execution sequence from the spatial pose of the soft robot. However, due to their complex structure, these models require large amounts of temporally continuous data \cite{zhou2022modeling}. In addition, most data-driven approaches focus solely on modeling the end of the soft robot, which is sufficient only for simple tasks. However, in complex tasks such as surgery, the soft robot is expected to avoid obstacles and ensure safe interaction with its environment, which necessitates modeling its entire body.

Soft robots are underactuated systems characterized by a complex mapping between input pressures and resulting morphology, making the design of control algorithms particularly challenging \cite{chenDataDrivenMethodsApplied2025}. Conventional control theory struggles to address whole-body control tasks for soft robots due to the high dimensionality of the state space. As an effective alternative, machine learning methods have been proposed to directly learn control policies, offering greater adaptability to complex and dynamic environments \cite{centurelliClosedLoopDynamicControl2022}. Researchers have employed deep Q-networks (DQN) \cite{satheeshbabu2019open} to develop open-loop position control strategies for pneumatic continuum robots, achieving target point tracking under quasi-static conditions. Other studies have combined trajectory optimization with supervised learning to derive closed-loop control policies \cite{thuruthelModelBasedReinforcementLearning2019}. Reinforcement learning has also demonstrated promising results in soft robot control \cite{zhouCableActuatedSoftManipulator2024b,  grauleSoMoGymToolkitDeveloping2022, berdicaReinforcementLearningControllers2024}. However, in real-world complex tasks such as surgery, soft robots are required to operate in black-box environments while ensuring safe interaction with surrounding tissues. This necessitates accurate whole-body modeling and control that comprehensively addresses the aforementioned challenges.

In this paper, we design a three-chamber pneumatic soft robot system for surgical applications. We propose a novel intelligent control framework for soft robot: the hysteresis-aware neural network modeling and whole-body reinforcement learning algorithm.

The main contributions of this paper can be summarized as follows.

\begin{itemize}
\item A soft robot system is designed for laparoscopic surgical operations. An accompanying soft robot control framework is proposed based on neural network dynamic modeling and reinforcement learning. Through the sim-to-real-to-sim process, consistency is ensured between simulation and real soft robots, bridging the sim-to-real gap to minimize performance drop.

\item A Hysteresis-aware Whole-Body Neural Network (HAW-NN) modeling approach is proposed, in which the direction of pneumatic actuation is incorporated to enable accurate prediction of hysteresis. The position and orientation of the entire soft robot are modeled in detail. This method is well-suited for scenarios that are challenging to model using conventional techniques, such as surgical soft robots composed of multiple materials or equipped with attachments.

\item Based on the Hysteresis-aware Whole-Body Neural Network (HAW-NN) model, we construct a highly parallel simulation environment for soft robot control and apply an on-policy reinforcement learning algorithm to efficiently train whole-body motion control policy. This approach provides a promising solution for complex operational tasks, such as surgical procedures.
\end{itemize}

\section{DESIGN AND FABRICATION OF SOFT ROBOTS}
To enhance the performance of the robot, we have designed a fiber-reinforced multidegree-of-freedom soft actuator, which effectively reduces the radial expansion of the actuator. The soft robot has a length of 70 mm and a diameter of 15 mm,  with a central 3 mm channel reserved for tools such as endoscopes and laser fibers. The air chambers are equally spaced at $120^{\circ}$ along the cylinder. The head and end segment are designed with embedded strong magnet caps (RbFeB magnet), enabling quick and flexible installation and detachment of the dual-segment actuator.

The fabrication of the robot mainly involves five main steps. First, assemble the molds and spray the release agent. Pour in liquid silicone and cure to create spiral pattern segments. Second, wrap the reinforcing fibers around the silicone in a spiral pattern with a symmetrical angle of 7 °. Third, after assembling the outer mold, pour liquid silicone to encapsulate the reinforcing fibers. Fourth, cast the head segment of the actuator. Fifth, cast the end segment and attach the air tubes, as shown in Fig.~\ref{fig:fabrication}.

\section{HYSTERESIS-AWARE WHOLE-BODY NEURAL NETWORK DYNAMIC MODELING OF SOFT ROBOTS}
Modeling soft robots presents several key challenges. Conventional analytical methods are limited in handling material inhomogeneity and integrated attachments. Data-driven approaches often fail to capture whole-body morphological information. Moreover, hysteresis effects induced by pressure history can result in significant positioning errors.

Conventional analytical models are generally applicable only to attachment-free soft robots. However, real-world applications often require robots with non-uniform and complex structures, such as those equipped with endoscopes or operative tools in surgical tasks. These additions lead to complex dynamic behaviors that cannot be accurately captured by traditional modeling methods.

In addition, soft robots exhibit noticeable hysteresis behavior, in which the robot position is influenced not only by current pressure values but also by the direction of pressure change. Our experiments on a three-chamber soft robot revealed that this hysteresis follows complex and often overlooked patterns, which can result in significant positioning errors. Specifically, when the same pressure is achieved through the pressurization and depressurization paths, the position of the end effector may differ significantly, with a maximum deviation of up to 3.4\% of the robot’s total length. Given the scale of the soft robot workspace, such deviations are non-negligible and must be accounted for in both modeling and control strategies.

To address these challenges, this paper proposes a hysteresis-aware neural network-based modeling approach that incorporates the hysteresis effect induced by the direction of actuation pressure and captures the whole-body configuration of the soft robot. Trained on real-world robot data, the model does not rely on predefined actuator structures or material-specific mechanics, thereby enabling accurate dynamic modeling of soft robots with complex structures.

\begin{figure}[!h]
    \centering
    \includegraphics[width=0.95\linewidth]{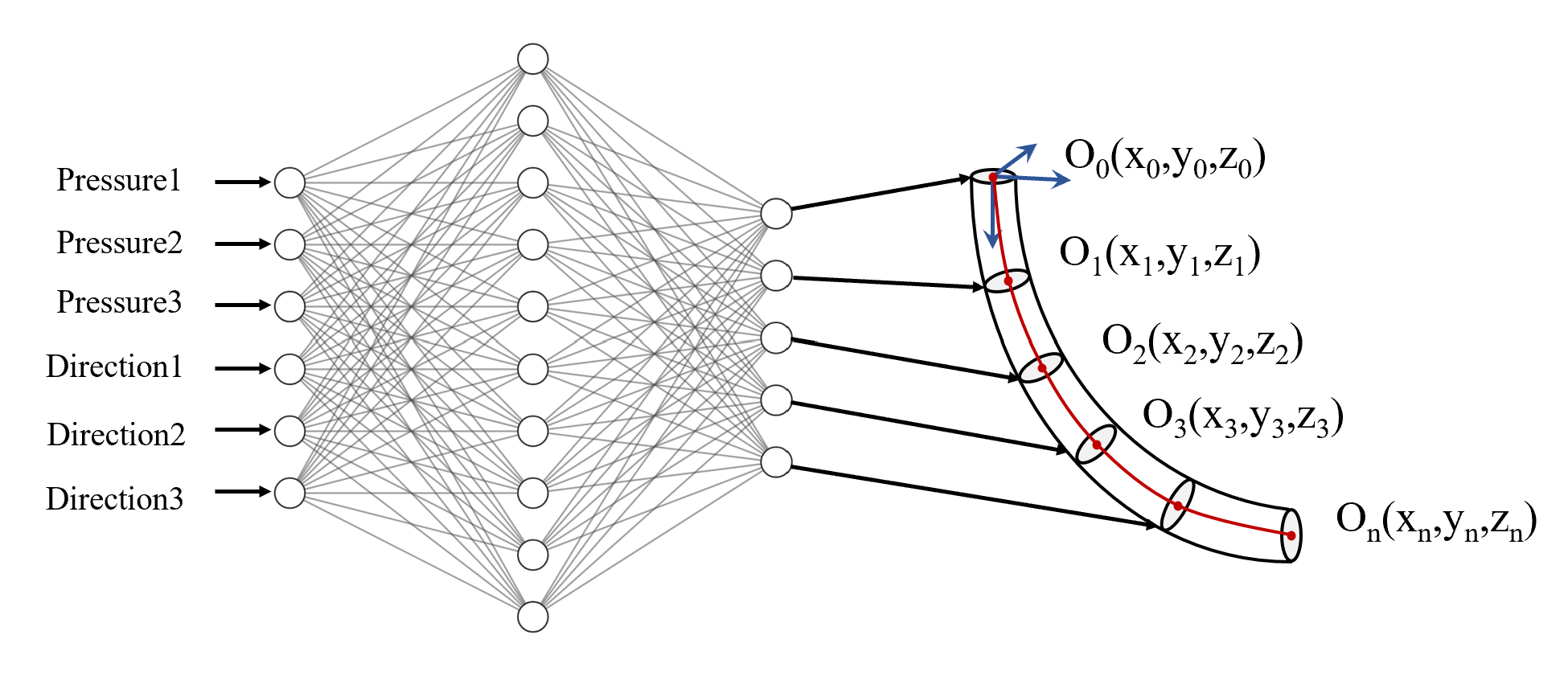}
    \caption{Schematic diagram of the Hysteresis-aware Whole-Body Neural Network (HWB-NN) model.}
    \label{fig:enter-label}
\end{figure}

\subsection{Hysteresis-aware Whole-Body Neural Network Model Design} 
The dynamic model of a soft robot describes the mapping between the input air pressure and the  morphological configuration the robot. The Piecewise Constant Curvature (PCC) model \cite{webster2010design} assumes that a continuum robot consists of a series of continuous circular arcs. Inspired by the PCC model, we propose a hysteresis-aware piecewise key feature points neural network model. In this approach, the soft body is represented as a series of connected segments, with key feature points between adjacent segments predicted by a neural network. The continuous morphology between these feature points is reconstructed using a B-spline curve fitting method. A circular cross-section perpendicular to the central axis of the soft robot is defined as a key feature plane. The center of mass of each key feature plane is designated as a key feature point $O_i(x_i, y_i, z_i)$. The vector $O_i - O_{i-1}$ captures the spatial continuity between adjacent key points. The modeling process only requires the ground truth coordinates of each piecewise key point. The proposed piecewise modeling method reduces the complexity of morphology data acquisition and simulation computation in 3D space, while providing a complete representation of the soft robot's whole-body morphological information. It also allows for adjustable complexity in morphology data acquisition and simulation computation based on task requirements, by varying the number of key feature points.

To address the strong nonlinearity and hysteresis in the relationship between actuation pressure and position, we propose a hysteresis-aware multilayer perceptron (MLP) architecture that incorporates both pressure magnitude and directional change as inputs. The specific network structure is as follows:

\begin{itemize}
\item Input layer: The input layer consists of 6 neurons, corresponding to the input air pressure values and their directions of change for the soft robot with 3 air chambers, and the input can be defined as:
\begin{align} 
 P = [p_1, p_2, p_3, d_1, d_2, d_3] \in \mathbb{R}^6.
\end{align}
\item Hidden layer: We experimentally evaluated the impact of different neural network architectures, varying both the number of hidden layers and the number of neurons per layer. The results of the comparative experiment are shown in Fig.~\ref{fig:model loss compare}. The model with the lowest test loss was selected, which consisted of 4 hidden layers, each containing 128 neurons.

\item Output layer: 
The output layer consists of $3\times n$ neurons corresponding to 3-dimensional coordinates of $n$ feature segmentation points, each of which can be represented as $O_i=(x_i,y_i,z_i)$ , so the output $Y$ is defined as:
\begin{align}
\mathbf Y &= [\mathbf O_1,..., \mathbf O_n]\in\mathbb{R}^{3\times n}\\
&=[O_{1x},O_{1y},..., O_{nx},O_{ny},O_{nz}] \nonumber.
\end{align}
\item Loss function:
The fixed and free ends of soft robots have different motion ranges that differ by several orders of magnitude. The naive MSE loss may introduce instability in model training. We proposed a weighted MSE based on the different motion ranges of feature points. The motion range is defined as:
\begin{align}
\Delta Y_{i}=max(Y_i)-min(Y_i).
\end{align}
The corresponding weights for each dimension output variable are defined as:
\begin{align}
w_i = 1+\frac{n\times \Delta Y_{i}}{\sum_{i=1}^{n} \Delta Y_{i}}.
\end{align}
The loss function weighted according to the range of motion is defined as:
\begin{align}
Loss(Y, \hat{Y}) = \frac{1}{n} \sum_{i=1}^{n} w_i \times (Y_i - \hat{Y}_i)^2.
\end{align}

\end{itemize}

During training, the model is evaluated on the validation set every 10 epochs. An early stopping criterion is applied such that if no improvement in validation loss is observed across 10 consecutive evaluations, the training process is terminated to prevent overfitting.

\subsection{Experimental Platforms Setup and Data Collection} 
\begin{figure}[h]
    \centering
    \includegraphics[width=1.0\linewidth]{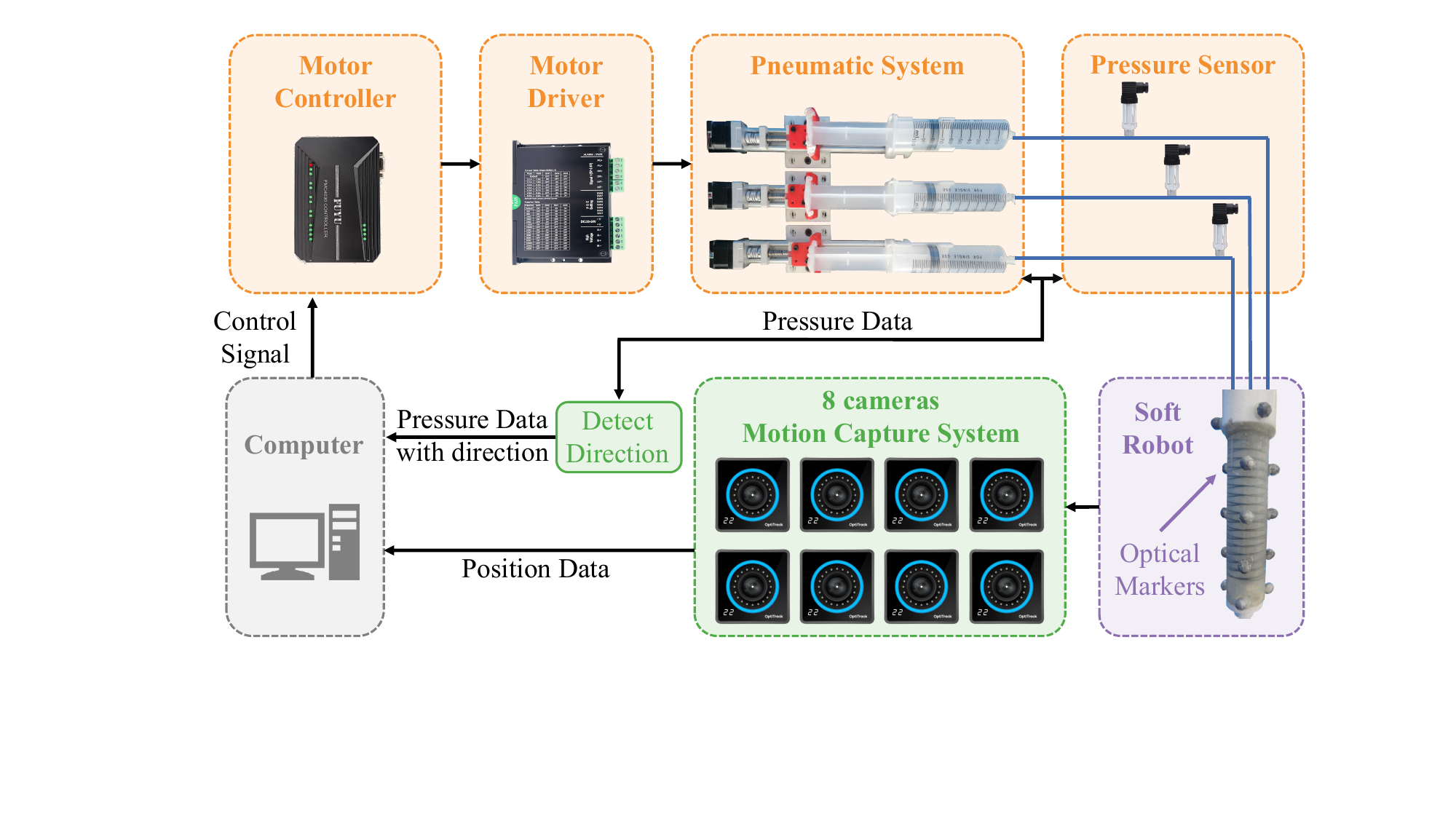}
    \caption{Schematic of the data acquisition system designed to capture directional pressure inputs and whole-body motion data of the soft robot.}
    \label{fig:data acquisition system}
\end{figure}

A schematic of the data acquisition system for the Hysteresis-aware Whole-Body Neural Network(HWB-NN) model is shown in Fig.~\ref {fig:data acquisition system}. The soft robot is fixed on the experimental table and driven by a pneumatic system consisting of a set of syringe pumps. Each chamber of the soft robot is equipped with a separate drive system. Each syringe pump is connected to an air chamber of the soft robot to form a closed air circuit. The piston position of the syringe pump is controlled by a lead screw slide table driven by a stepper motor, and the air pressure in the chamber is controlled by changing the volume in the syringe pump. The advantage of this air pressure control system is its ability to provide negative pressure in the soft body chamber. A pneumatic pressure sensor is used to measure the air pressure in the pneumatic system in real-time, which is sent back to the computer as feedback for proportional-integral-derivative (PID) control. The PID algorithm calculates a control signal for the motor speed based on the desired air pressure and sends it to the motor controller to execute. The soft robot generates different deformation motions according to the air pressure in the air chamber.

The experimental table of soft robot is surrounded by an infrared optical motion capture system composed of eight OptiTrack cameras that collect millimeter-level position data. 
The cameras are spaced to capture the markers on the soft robot despite its bending. We selected $n$ planes perpendicular to the axis direction of the soft robot as the key feature planes. The number of planes $n$ is determined according to the modeling resolution requirement. 
To ensure robust capture of the deformation of soft robots during bending and elongation, at least 3 infrared reflective markers are affixed to each key feature plane. 
These markers enable accurate estimation of the plane center using a circle fitting algorithm. 
\vspace{-3pt}
Markers should be arranged between the air chambers, rather than close to the outside of the air chambers, to minimize the effect of the transverse expansion of the air chambers on the circle fitting. The first circle marker is attached to the fixed bracket at the end of the soft robot to obtain the initial point coordinates.

To collect a realistic kinematic dataset of the soft robot, the pressure in each air chamber is sequentially varied to traverse the entire action space at defined step intervals. Data collection was conducted under quasi-static conditions. After each controlled set of desired pressures is applied, a 3-second delay is introduced to allow the soft robot to stabilize. Subsequently, 1 second of data is recorded, including air pressure readings from pneumatic pressure sensors and positional data from markers located on the surface of soft robot, captured by the motion capture system. The positional data of the markers are processed using a circle fitting algorithm to determine the center coordinates of each piecewise key feature plane, which are used as key feature points. Data are collected at a frequency of 20 Hz, and the mean value of each 1-second data segment is computed to represent a single valid data point. The input of the dataset consists of three-dimensional pressure values and their corresponding directional changes in each dimension. The directions of pressure increase and decrease are represented by 1 and -1, respectively. The output includes the spatial coordinates of the piecewise key feature points. A total of 13824 samples were collected as the training set. Additionally, validation and test sets containing 1000 samples each were collected separately, with distributions different from the training set.

\section{REINFORCEMENT LEARNING WHOLE-BODY CONTROL FOR SOFT ROBOTS}
To solve the problem of high-dimensional state space in soft robot whole-body control, address complex task constraints, and narrow the sim-to-real gap, we propose a reinforcement learning method based on a parallel simulation environment built on the Hysteresis-aware Whole-Body Neural Network (HAW-NN) model. 
We first modeled the control process as a Markov Decision Process (MDP) \cite{puterman1990markov}. The policy is then trained using the Proximal Policy Optimization (PPO) \cite{schulman2017proximal} algorithm. The implementation details are discussed in detail below.

\subsection{{MDP-Based Modeling of the Robot System}}
To apply reinforcement learning within a soft robot system, the motion process of the soft robot is represented as an MDP to capture the interaction between the robot and its environment. 
The Markov Decision Process is defined by five components:
\begin{align}
(S,A,P_a,R_a,\gamma),
\end{align}
where $S$ represents the state space, $A$ denotes the action space, $P_a$ is the transition function, $R_a$ is the reward function, and $\gamma$ is the discount factor.

$S$ is the state space, the coordinates of the whole body position of the soft robot are available in our proposed whole-body neural network model, so the contents of the state space can be specifically adjusted according to the task requirements. 
For the tracking task, where each state $s$ consists of: the robot position coordinates $Y_t^{body}\in\mathbb{R}^{3n}$ ($n$ is the number of piecewise key feature points on the whole body of the soft robot), the current tracking error $e_t=Y^{tar}-Y_t^{body}\in\mathbb{R}^{3n}$ ( $Y^{tar}$ denotes the target whole-body key feature point coordinates), air pressure state for the three chambers (with values and directions) $P_t\in\mathbb{R}^6$, and time $t$. The state is hence defined as:
\begin{align}
s_t=[Y_t^{body},e_t,P_t,t].
\end{align}

$A$ is the action space, and the action $a_t \in A$ at timestamp $t$ is defined as the change in pressure value $\Delta p\in\mathbb{R}^3$ in each interaction step, which is set to be a continuous value in the range of -3~kPa to 3~kPa.
\begin{align}
a_t=\Delta p\in\mathbb{R}^3
\end{align}
$P_a(s,s^{\prime})$ is the state transition function, representing the probability that taking action aa will cause the system to transition from state $s$ to state $s'$. In this work, the transition dynamics are defined based on the HWB-NN model:
\begin{align}
s_{t+1}=T_{nn}(s_t,a_t).
\end{align}
$R_a(s,s^\prime)$ is the reward function, which represents the real-time expected reward corresponding to the transfer from state $S$ to state $S'$, and is used to assess whether the agent's behavior in performing the task is optimized towards the goal. For example, in our tracking task, the reward is defined as the exponent of negative Euclidean distance error in 3D space:
\begin{align}
r_t=\exp(-\|Y^{\text{tar}} - Y^{\text{body}}\|).
\end{align}

\subsection{{Policy Training with Proximal Policy Optimization}}
Second-order optimization methods have been widely used in reinforcement learning due to their ability to provide more stable and theoretically grounded policy updates. Methods such as Trust Region Policy Optimization (TRPO) \cite{schulman2015trust} involve the computation or approximation of second-order derivatives, such as the Hessian or the Fisher Information Matrix. These operations are computationally intensive and require solving large linear systems or performing matrix inversions, making them inefficient in practice, especially in high-dimensional policy spaces such as whole-body control tasks in soft robots.
To address these computational challenges, we adopt Proximal Policy Optimization (PPO), a first-order optimization algorithm that avoids the need for complex second-order computations. PPO directly optimizes the policy by using a clipped surrogate objective, which constrains the policy update within a trust region-like bound. This ensures training stability while maintaining computational efficiency, making PPO well-suited for complex tasks like soft robot whole-body motion control.
PPO optimizes a clipped surrogate objective function, designed to limit the deviation between new and old policies. The objective function is defined as follows:
\begin{align}
 & L^{\text{CLIP}}(\theta) \\
 & =\mathbb{E}_t\left[\min\left(r_t(\theta)\hat{A}_t,\mathrm{clip}(r_t(\theta),1-\epsilon,1+\epsilon)\hat{A}_t\right)\right]\nonumber,
\end{align}
where $r_t(\theta)=\frac{\pi_\theta(a_t|s_t)}{\pi_{\theta_{\mathrm{old}}}(a_t|s_t)}$ is the probability ratio, $\hat{A}_t$ is the advantage estimate at time $t$, and $\epsilon$ is a hyperparameter that controls the trade-off between policy flexibility and update stability.

The policy is represented by an MLP architecture, consisting of 2 hidden layers with 64 units each and a hyperbolic tangent (Tanh) activation function. The actor and critic share the same feature extraction layers, followed by separate output heads for action distribution and state value estimation, respectively.

The reinforcement learning environment is constructed based on a neural network-based kinematic model that incorporates hysteresis-aware modeling. At each time step, the environment receives an action input $\Delta p$, representing the change in pressure for each pneumatic channel. The direction of pressure variation is detected from the sign of $\Delta p$, which is essential for capture the pressure-dependent hysteresis behavior. Based on the magnitude and direction of the pressure in the current step, the neural network model accurately predicts the whole-body position of the soft robot. In addition, the environment receives a predefined target state of the soft robot and can incorporate positional or angular constraints on specific body parts. Based on these inputs, it computes a reward signal at each step. The reinforcement learning agent generates actions based on current observation to optimize task performance.

\subsection{{Implementation Details}}

\paragraph {Action PID controller}
The action output from the policy network is sent to a PID controller, which regulates the internal pressure of the soft robot to match the target pressure specified by the action. The output of the PID controller determines the actuation speed of a motor that drives the syringe connected to the pneumatic chamber.
Considering the high inertia of the soft robotic system, the PID parameters are carefully tuned. The proportional gain $K_p$ is set to a relatively low value to avoid overshoot and oscillations. A relatively small value is assigned to the integral gain $K_i$ to prevent excessive error accumulation. A moderate derivative gain $K_d$ is selected to anticipate pressure trends and mitigate overshoot. 

\paragraph{Parallel Environments} 
To accelerate training efficiency, multiple environments are executed in parallel. A larger number of parallel environments generally leads to faster policy updates and improved sample diversity, as long as the available GPU memory allows. In our experiments, we used an NVIDIA RTX 4090 GPU and set the number of parallel environments to 64.

\paragraph{Hyperparameters}
The configuration of hyperparameters used in training is listed in Table ~\ref{tab:hyperparams}.
\begin{table}[h]
\centering
\caption{PPO Hyperparameters}
\begin{tabular}{cc}
\toprule
\textbf{Hyperparameter} & \textbf{Value} \\
\midrule
Discount Factor, $\gamma$ & 0.99 \\
GAE, $\lambda$ & 0.95 \\
Entropy Coefficient & 0.01 \\
Clip Range & 0.2\\
Batch Size & 4096\\
Number of training steps & 20,000,000 \\
\bottomrule
\end{tabular}
\label{tab:hyperparams}
\end{table}
\section{EXPERIMENTS}
To evaluate the advantages of the proposed Hysteresis-aware Whole-Body Neural Network (HWB-NN) model over conventional method, we conducted accuracy comparison experiments on a test data set.
Furthermore, to assess the effectiveness of the proposed reinforcement learning framework in real-world control, the learned policy is deployed on the real soft robot and evaluated through a trajectory tracking task.
To demonstrate the effectiveness of the system in controlling multiple soft robots in complex tasks, we select a representative endoscopic surgery scenario, a typical application setting for soft robots, and conduct experiments using real robots operating on a surgical phantom.
\begin{figure}[!h]
    \centering
    \includegraphics[width=1.0\linewidth]{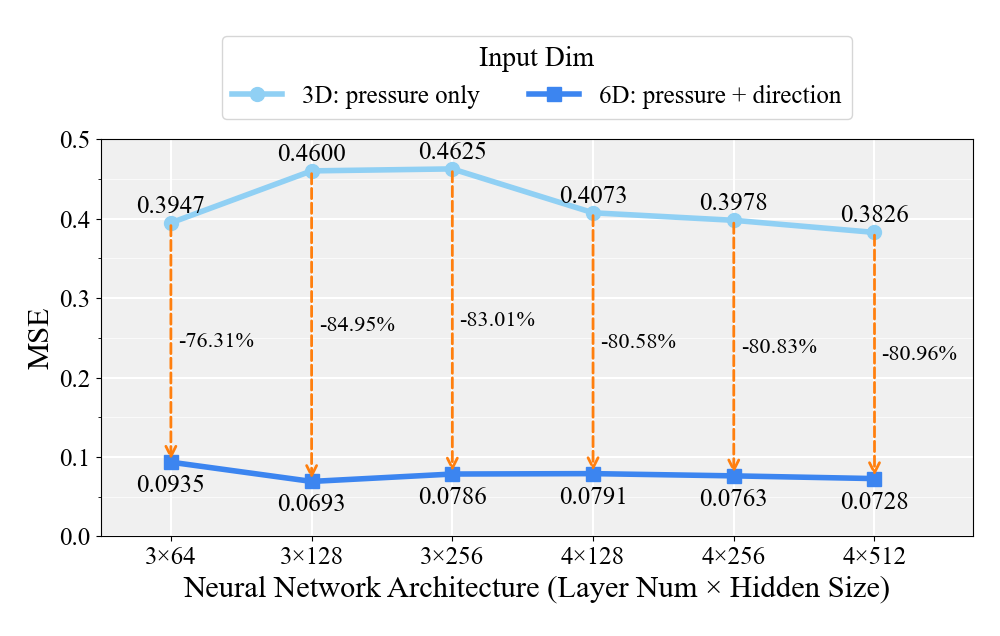}
    \caption{Comparison of kinematic modeling performance across six neural network architectures under two input settings: 3D pressure-only and 6D pressure-plus-direction (proposed). Mean squared error (MSE) is used as the evaluation metric.}
    \label{fig:model loss compare}
\end{figure}
\subsection{Modeling Results and Analysis}
To validate the effectiveness of the proposed Hysteresis-aware Whole-Body Neural Network(HWB-NN) model, two types of networks were trained on the same test dataset, consisting of 1,000 points distributed throughout the action space. One type takes the proposed 6-dimensional input that includes both pressure values and their directional changes, while the other uses only 3 dimension pressure magnitudes, which is a conventional approach. The resulting neural network dynamic models were evaluated on the test set using Mean Squared Error (MSE) to assess prediction accuracy. In addition, 6 different neural network architectures were compared to examine the impact of model structure on kinematic modeling performance. As shown in Fig.~\ref{fig:model loss compare}, the proposed method that incorporates pressure direction achieves significantly better accuracy than the conventional approach that considers only pressure magnitudes. For networks with 3 to 4 layers, the size and architecture of the network had relatively minor effects on prediction error.
Across the six evaluated network architectures, the lowest MSE loss obtained using the 3-dimensional input model was 0.3947, while the 6-dimensional input model reached a minimum loss of 0.0693, highlighting the superior modeling capability and effectiveness of the proposed hysteresis-aware framework.

\subsection{Simulation Results}
\begin{figure}[!h]
    \centering
    \begin{subfigure}[b]{0.45\linewidth}
        \centering
        \includegraphics[width=\linewidth]{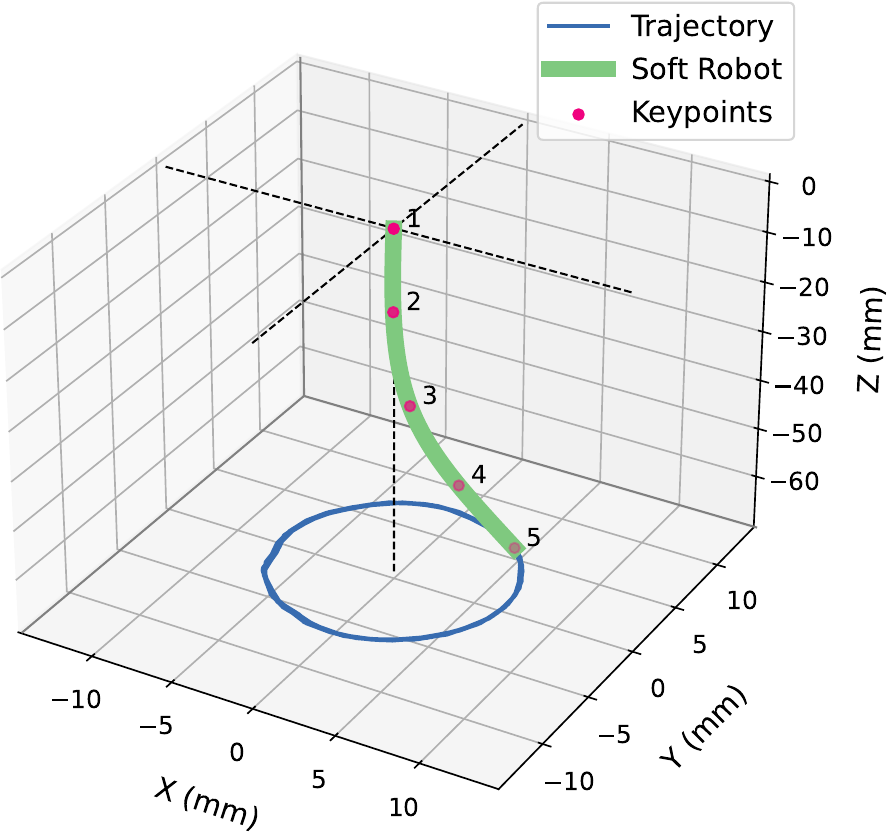}
        \caption{Circle trajectory}
        \label{fig:circle}
    \end{subfigure}
    \begin{subfigure}[b]{0.45\linewidth}
        \centering
        \includegraphics[width=\linewidth]{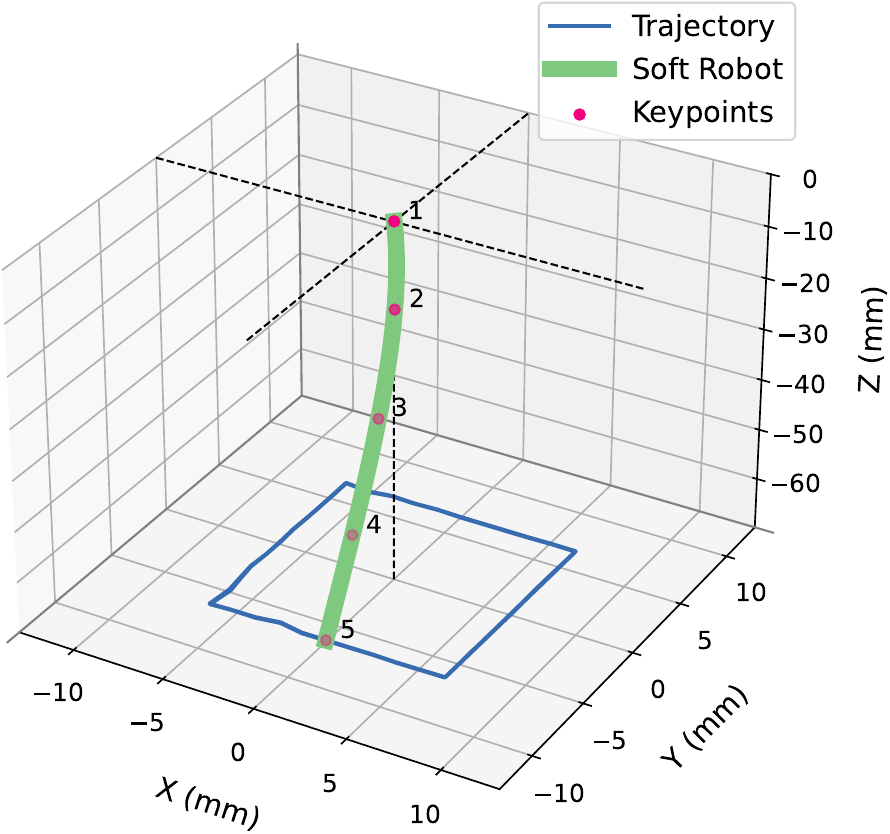}
        \caption{Square trajectory}
        \label{fig:sq13}
    \end{subfigure}
    \caption{ Simulation result of square trajectory tracking by the soft robot.}
    \label{fig:sim trajectory compare}
\end{figure}

The trained reinforcement learning policy is evaluated in simulation by performing a trajectory tracking task, where the soft robot is required to track a target trajectory with its end effector. The tracking performance is shown in Fig.~\ref{fig:sim trajectory compare} . The blue line indicates the end effector’s motion path in simulation, while the green curve shows the predicted whole-body configuration of the robot. Five key feature points along the body are highlighted in magenta, demonstrating the availability of full-body pose information during tracking.
\begin{figure}[!h]
    \centering
    \begin{subfigure}[b]{0.45\linewidth}
        \centering
        \includegraphics[width=\linewidth]{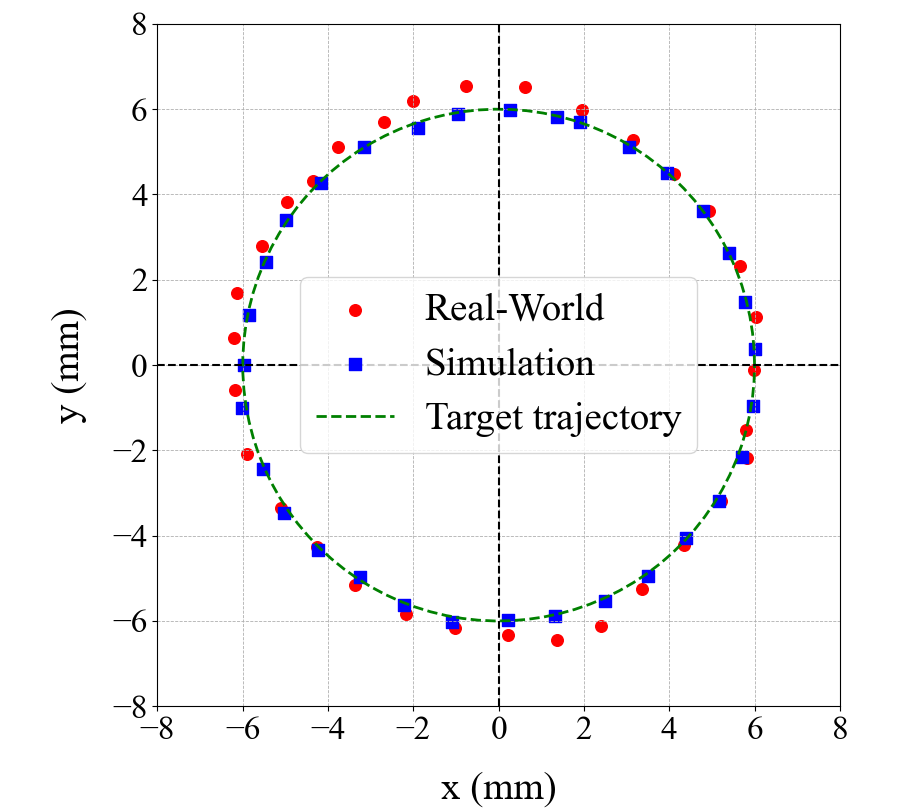}
        \caption{Circle trajectory}
        \label{fig:circle}
    \end{subfigure}
    \begin{subfigure}[b]{0.45\linewidth}
        \centering
        \includegraphics[width=\linewidth]{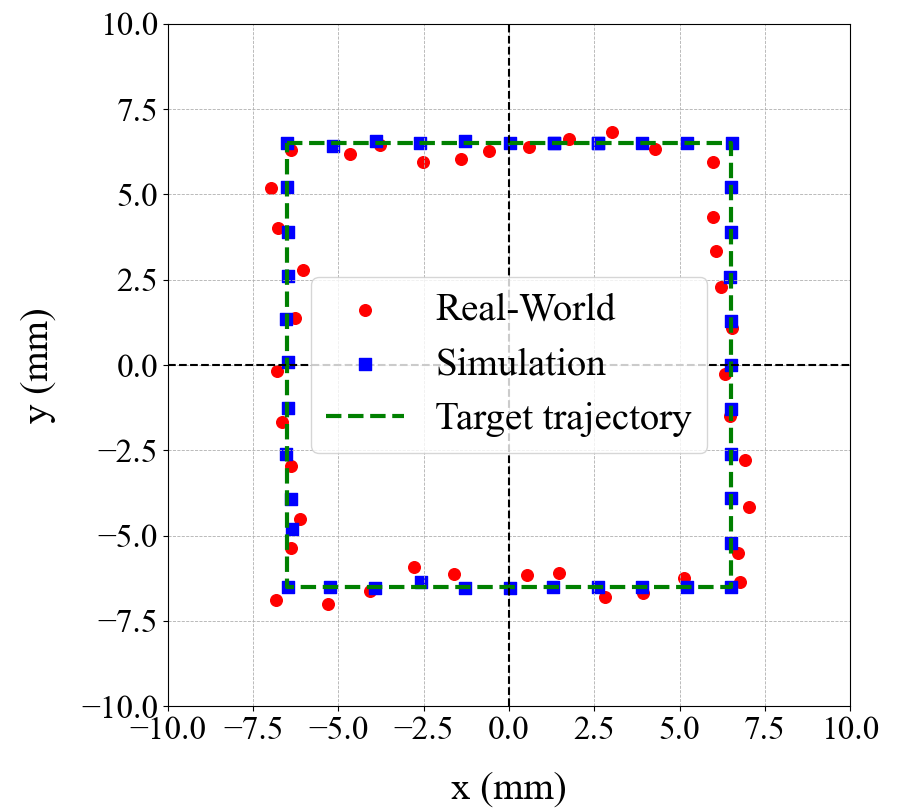}
        \caption{Square trajectory}
        \label{fig:sq13}
    \end{subfigure}
    \caption{ Tracking results of the soft robot on (a) circular and (b) square trajectories in simulation and real-world experiments.}
    \label{fig:sim real trajectory compare}
\end{figure}

\begin{table}[!h]
\centering
\caption{MSEs in Simulation and Real-World Experiments Relative to Target Trajectories}
\begin{tabular}{ccc}
\toprule
\textbf{Trajectory} &\textbf{Simulation} &\textbf{Real-World} \\
\midrule
Circle & 0.0438 mm & 0.250 mm \\
Square & 0.0104 mm & 0.126 mm \\
\bottomrule
\end{tabular}
\label{tab:tracking error}
\end{table}

\subsection{Real-World Soft Robot Deployment Results}
To evaluate the effectiveness of the learned control policy in the actual soft robot, we deployed the trained policy on the real system and recorded the movements using a motion capture system. The tracked trajectories in the real world were compared with those obtained in simulation, as illustrated in Fig.~\ref{fig:sim real trajectory compare}, where the trajectories are projected onto the X-Y plane for visualization. 
\renewcommand{\thefigure}{9}
\begin{figure*}[!b]
    \centering
    \includegraphics[width=0.9\linewidth]{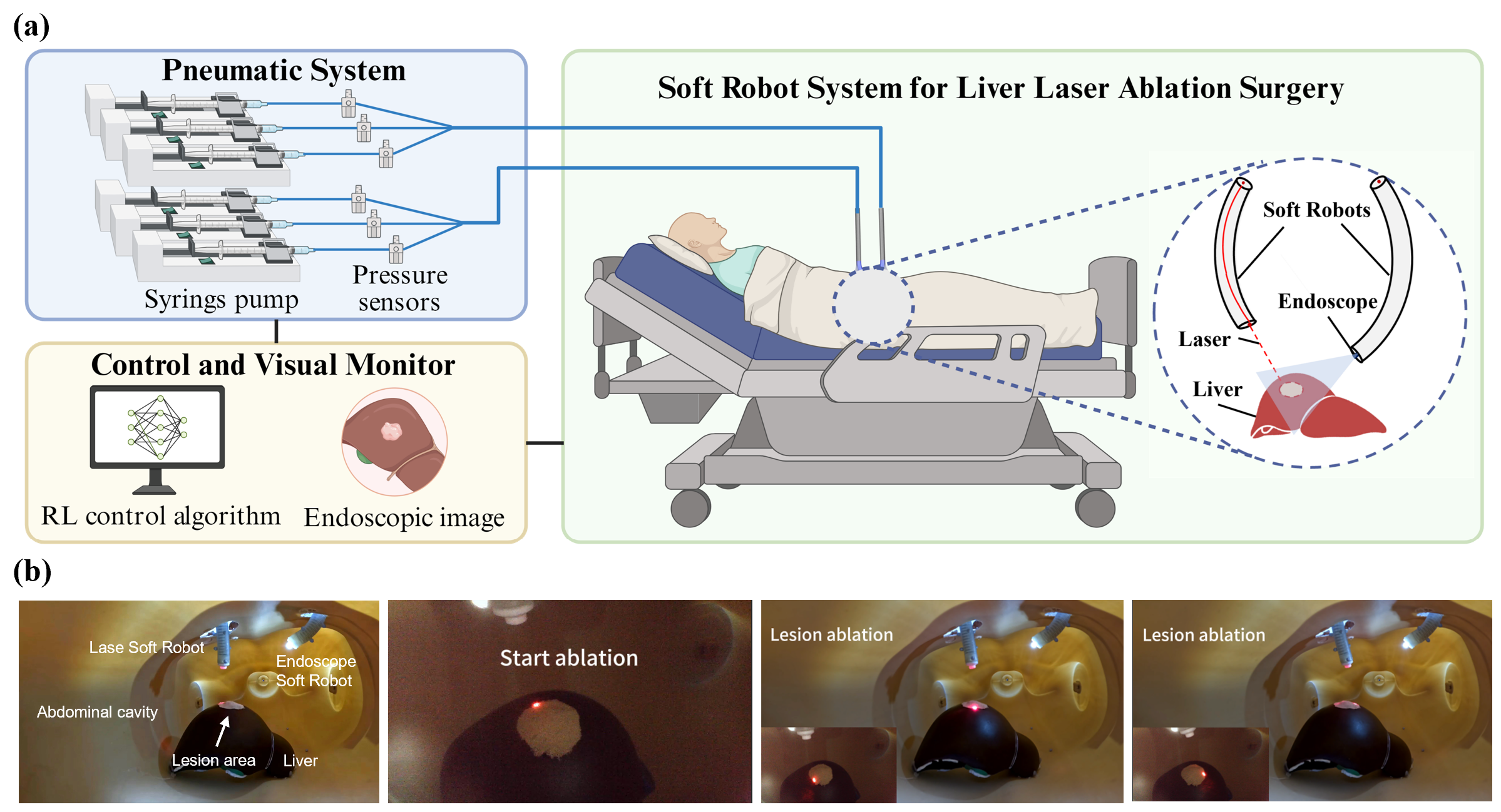}
    \caption{(a) Overview of the multiple soft robot system for laparoscopic surgery, demonstrating coordinated control and task execution. (b) Time-sequenced snapshots of the liver laser ablation procedure.} 
    \label{fig:surgery system and snapshots}
\end{figure*}
\setcounter{figure}{7}
\renewcommand{\thefigure}
{\arabic{figure}}

The average Euclidean errors for tracking circular and square trajectories in simulation and real-world experiments are summarized in Table~\ref{tab:tracking error}.

\subsection{Complex Task Whole-Body Control Experiments}
To further demonstrate the advantages of our whole-body information based control approach in complex tasks, we design a representative application task for laparoscopic surgery. In this scenario, a laser is mounted in the center of the soft robot, and the objective is to control the robot such that the laser beam intersects a plane located at a distance and follows a predefined trajectory to ablate pathological tissues.
\begin{figure}[!h]
    \centering
    \begin{subfigure}[b]{0.47\linewidth}
        \centering
        \includegraphics[width=\linewidth]{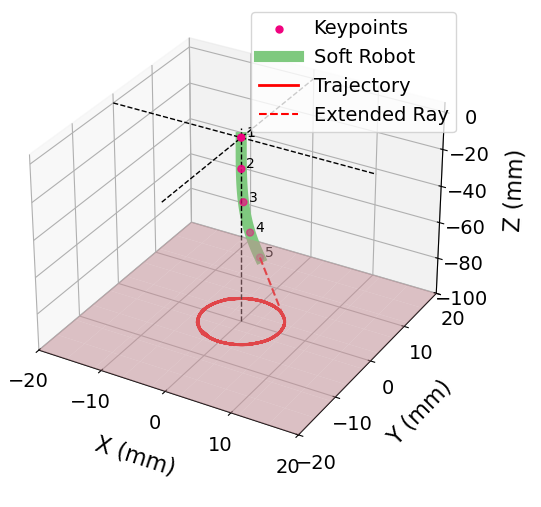}
        \caption{Simulation}
        \label{fig:circle}
    \end{subfigure}
    \hfill
    \begin{subfigure}[b]{0.42\linewidth}
        \centering
        \includegraphics[width=\linewidth]{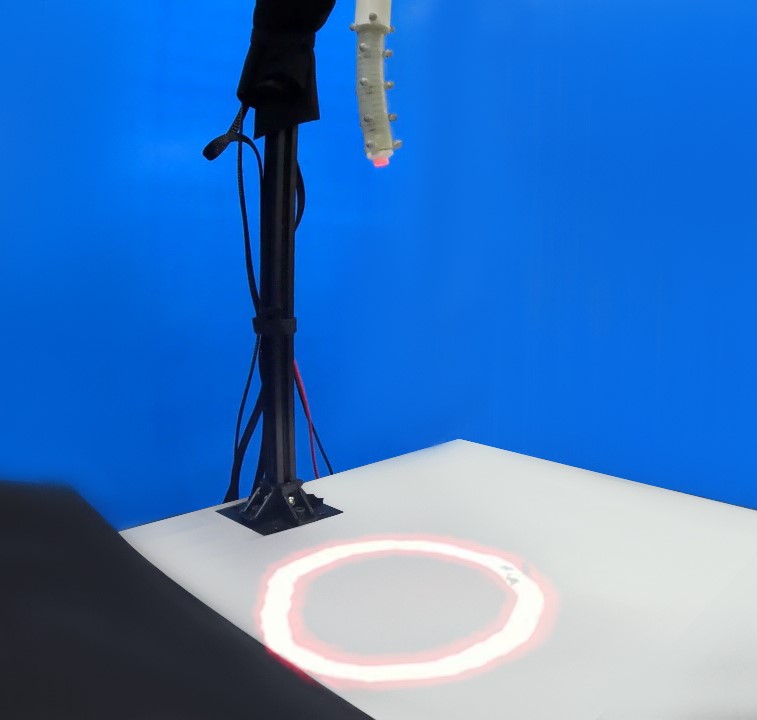}
        \caption{Real-World}
        \label{fig:sq13}
    \end{subfigure}
    \caption{Laser pointing experiment to validate whole-body control.}
    \label{fig:laser}
\end{figure}
This task requires accurate estimation of the laser pointing direction, which depends on the global curvature of the soft robot body. Conventional modeling and control methods are insufficient to handle this due to their limited capacity in capturing whole-body deformation. Using our hysteresis-aware whole-body modeling approach, the simulation environment can accurately predict the laser projection point on the remote plane. Combined with the PPO reinforcement learning algorithm, which is well-suited for handling high-dimensional state spaces, the framework can effectively learn control policies to accomplish this complex task. The average Eulerian error between the intersection trajectory and the target trajectory in the simulation experiments is 0.0419~mm, as shown in Fig.~\ref{fig:laser}.

To evaluate the effectiveness of coordinated control for multiple soft robots, we conducted a simulated liver laser ablation surgery experiment using a mannequin model. The task was collaboratively performed by two soft robots: one equipped with a laser module and the other with an endoscope. The endoscope-equipped soft robot first bent toward the target position to provide video monitoring of the surgical site. Subsequently, the laser-equipped soft robot was used to perform circular ablation cutting around the lesion area, as shown in Fig.~\ref{fig:surgery system and snapshots}. 
The laser-equipped soft robot accurately completed a full circular ablation around a 20~mm-diameter lesion within 60 seconds.
The experimental results demonstrate that the proposed control method achieves high accuracy, stability, and coordination in the collaborative control of multiple robots.
\section{CONCLUSIONS}
This paper presents a reinforcement learning control approach for soft robots based on a Hysteresis-aware Whole-Body Neural Network(HWB-NN) model. To address tasks with morphological constraints, a hysteresis-aware piecewise key feature point neural network model is proposed, which represents the full-body state of the soft robot by modeling a limited set of key feature points. To account for the hysteresis characteristics of soft robots, the proposed neural network model integrates pressure direction information, which effectively reduces the prediction errors associated with historical pressure paths. To achieve whole-body control with high-dimensional state space, we construct a parallelized simulation environment based on the accurate HWB-NN model and apply the reinforcement learning algorithm to efficiently train the control policy. This approach offers a promising solution for complex operational tasks, such as surgical procedures.

The proposed data-driven modeling approach reduces the simulation-to-reality gap and is applicable to soft robots composed of various materials or complex structural combinations. However, achieving high prediction accuracy relies on large volumes of training data. In future work, we aim to improve data utilization and develop more efficient data collection strategies. We also plan to explore temporal modeling approaches, such as Long Short-Term Memory (LSTM) networks, to strengthen the dynamic modeling capabilities and further improve the generalization performance of the learned model to real-world control tasks.





\section*{ACKNOWLEDGMENT}
This research was supported by National Natural Science Foundation of China (No.62406159, 62325405, 62403307), Postdoctoral Fellowship Program of CPSF under Grant Number (GZC20240830, 2024M761676), China Postdoctoral Science Special Foundation 2024T170496, the Science and Technology Commission of Shanghai Municipality (24ZR1439800). 

Fig.~9(a) was created with BioRender.com and is used with permission under a BioRender Publication License.





\bibliographystyle{ieeetr} 
\bibliography{filtered_softRL2} 

\end{document}